\pdfoutput=1
\def\year{2021}\relax

\documentclass[letterpaper]{article} 
\usepackage{aaai21} 
\usepackage{times} 
\usepackage{helvet} 
\usepackage{courier} 
\usepackage[hyphens]{url}
\usepackage{graphicx} 
\usepackage{graphicx}
\usepackage{natbib}
\usepackage{caption}
\usepackage[T1]{fontenc}    
\usepackage{url}             
\usepackage{booktabs}       
\usepackage{amsfonts}        
\usepackage{nicefrac}        
\usepackage{microtype}       
\usepackage{subcaption}
\usepackage{enumitem}
\usepackage{amsmath}       
\usepackage{amssymb}
\usepackage{tablefootnote}
\usepackage{multirow}
\usepackage{array}

\usepackage{hyperref}
\usepackage{lscape}
\usepackage{multicol}
\usepackage{stfloats}
\usepackage{xcolor}
\usepackage[switch]{lineno}
\usepackage{makecell}

\DeclareMathOperator{\sign}{sign}
\DeclareMathOperator{\sech}{sech}

\title{A Neuro-Inspired Autoencoding Defense Against Adversarial Perturbations}

\author{
  Can Bakiskan, Metehan Cekic, Ahmet Dundar Sezer, Upamanyu Madhow  \\
}
\affiliations {
  Department of Electrical and Computer Engineering\\
  University of California, Santa Barbara\\
  Santa Barbara, CA 93106 \\
  \texttt{\{canbakiskan,metehancekic,adsezer,madhow\}@ece.ucsb.edu} \\
}

\begin{document}
\maketitle

\begin{abstract}
Deep Neural Networks (DNNs) are vulnerable to adversarial attacks: carefully constructed perturbations to an image can seriously impair classification accuracy, while being imperceptible to humans. While there has been a significant amount of research on defending against such attacks, most defenses based on systematic design principles have been defeated by appropriately modified attacks. For a fixed set of data, the most effective current defense is to train the network using adversarially perturbed examples. In this paper, we investigate a radically different, neuro-inspired defense mechanism, starting from the observation that human vision is virtually unaffected by adversarial examples designed for machines. We aim to reject $\ell^{\infty}$ bounded adversarial perturbations {\it before} they reach a classifier DNN, using an encoder with characteristics commonly observed in biological vision: sparse overcomplete representations, randomness due to synaptic noise, and drastic nonlinearities. Encoder training is unsupervised, using standard dictionary learning.  A CNN-based decoder restores the size of the encoder output to that of the original image, enabling the use of a standard CNN for classification. Our nominal design is to train the decoder and classifier together in standard supervised fashion, but we also consider unsupervised decoder training based on a regression objective (as in a conventional autoencoder) with separate supervised training of the classifier.  Unlike adversarial training, all training is based on {\it clean} images.

Our experiments on the CIFAR-10 show performance competitive with state-of-the-art defenses based on adversarial training, and point to the promise of neuro-inspired techniques for the design of robust neural networks. In addition, we provide results for a subset of the Imagenet dataset to verify that our approach scales to larger images.
\end{abstract}


\section{Introduction} \label{sec:intro}

The susceptibility of neural networks to small, carefully crafted input perturbations raises great concern regarding their robustness and security, despite their immense success in a wide variety of fields: computer vision \cite{he2015delving,7913730}, game playing agents \cite{silver2017mastering}, and natural language processing \cite{vaswani2017attention}. 

Since this vulnerability of DNNs was pointed out \cite{biggio2013evasion,szegedy2013intriguing,goodfellow2014adversarial}, there have been numerous studies on how to generate these perturbations (adversarial attacks) \cite{goodfellow2014adversarial,kurakin2016physical,carlini2016distillationrefuted,madry2017towards} and how to defend against them \cite{madry2017towards,wong2017provable,guo2017countering,yang2019me,buckman2018thermometer}. 
Existing defenses that attempt to employ systematic or provable techniques either do not scale to large networks \cite{wong2017provable}, or have been defeated by appropriately modified attacks \cite{guo2017countering,yang2019me,buckman2018thermometer}. State of the art defenses \cite{madry2017towards,zhang2019theoretically,carmon2019unlabeled} employ adversarial training (i.e., training the model with adversarially perturbed examples), but
there is little insight into how DNNs designed in this end-to-end, ``top down'' fashion provide robust performance, and how they might perform against a yet-to-be-devised attack.
Classification performance with attacked images is still well below that with clean images, hence there remain fundamental security concerns as we seek to deploy DNNs in safety-critical applications such as vehicular automation, in addition to standard concerns regarding inference for tail events not seen during training.

\begin{figure}[t]
    \centering
        \includegraphics[width=\linewidth]{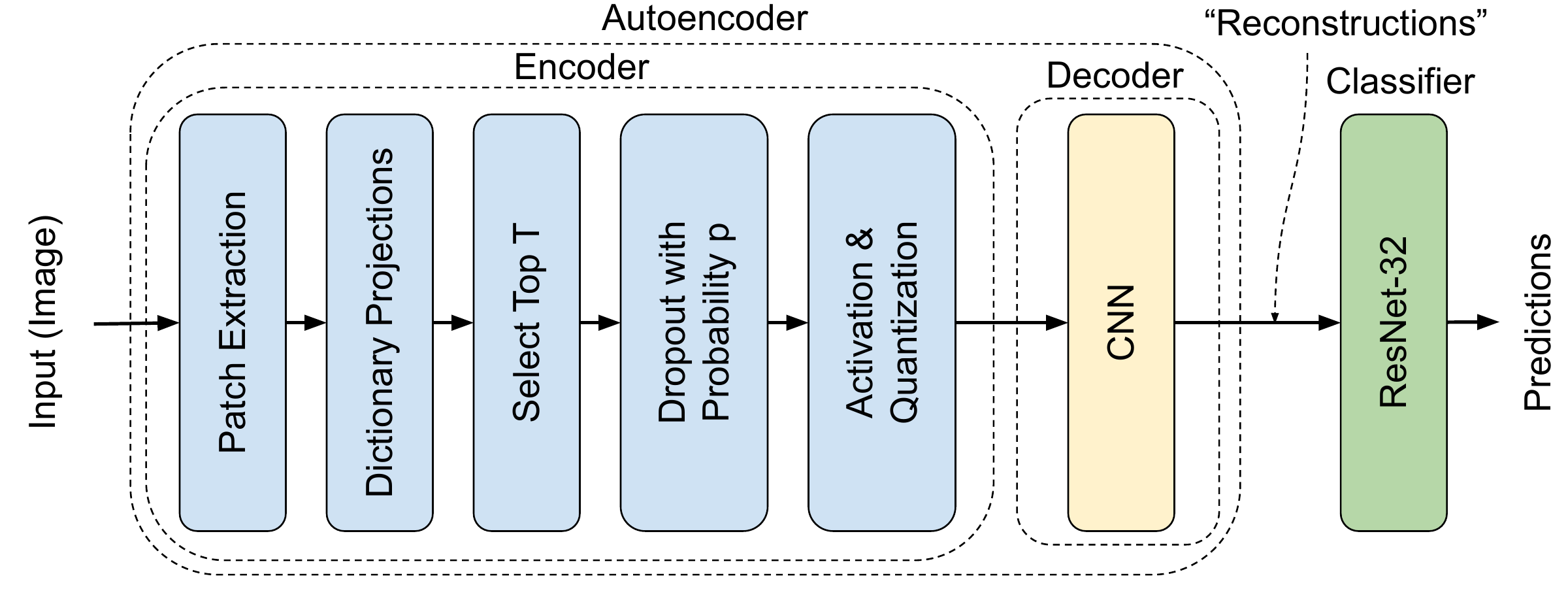}
    \caption{Proposed autoencoding defense. Decoder restores input size but does not attempt to reconstruct the input in our nominal design (supervised decoder+classifier training).}
    \label{fig:block_diagram}
\end{figure}

\noindent
{\bf Approach:} In this paper, we turn to neuro-inspiration for design insights for defending against adversarial attacks, inspired by the observation that humans barely register adversarial perturbations devised for machines.  While neuro-inspiration could ultimately provide a general framework for designing DNNs which are robust to a variety of perturbations, in this paper, we take a first step by focusing on the well-known $\ell^\infty$ bounded attack, which captures the concept of ``barely noticeable'' perturbation. Our architecture, illustrated in Figure~\ref{fig:block_diagram}, does not require adversarial training: it consists of (a) a neuro-inspired encoder which is learnt in a purely unsupervised manner, (b) a decoder which produces an output of the same size as the original image, (c) a standard CNN for classification.
The decoder and classifier are trained in standard supervised fashion using {\it clean} images passed through our encoder. The key features we incorporate into our encoder design are sparsity and overcompleteness, long conjectured to be characteristic of the visual system \cite{olshausen1997sparse}, lateral inhibition \cite{blakemore1970lateral}, synaptic noise  \cite{prescott2003gain,pattadkal2018emergent}, and drastic nonlinearity \cite{prenger2004nonlinear}.

We use standard unsupervised dictionary learning \cite{mairal2009online} to learn a sparse, highly overcomplete (5-10X relative to ambient dimension) patch-level representations. However, we use the learnt dictionary in a non-standard manner in the encoder, not attempting patch-level reconstructions. Specifically, we take the top $T$ coefficients from each patch (lateral inhibition), randomly drop a fraction $p$ of them (synaptic noise and lateral inhibition), and threshold and quantize them, retaining only their sign (drastic nonlinearity).  This encoder design is the key step in attenuating adversarial perturbations, as we show via analysis of the empirical statistics of the encoder outputs. We use overlapping patches (providing an additional degree of overcompleteness). The patch-level outputs, which have ternary quantized entries, 
are fed to a multi-layer CNN decoder whose output is the same size as the original RGB image input. This is then fed to a standard classifier DNN.

\begin{figure}[t]
    \centering
        \includegraphics[width=\linewidth]{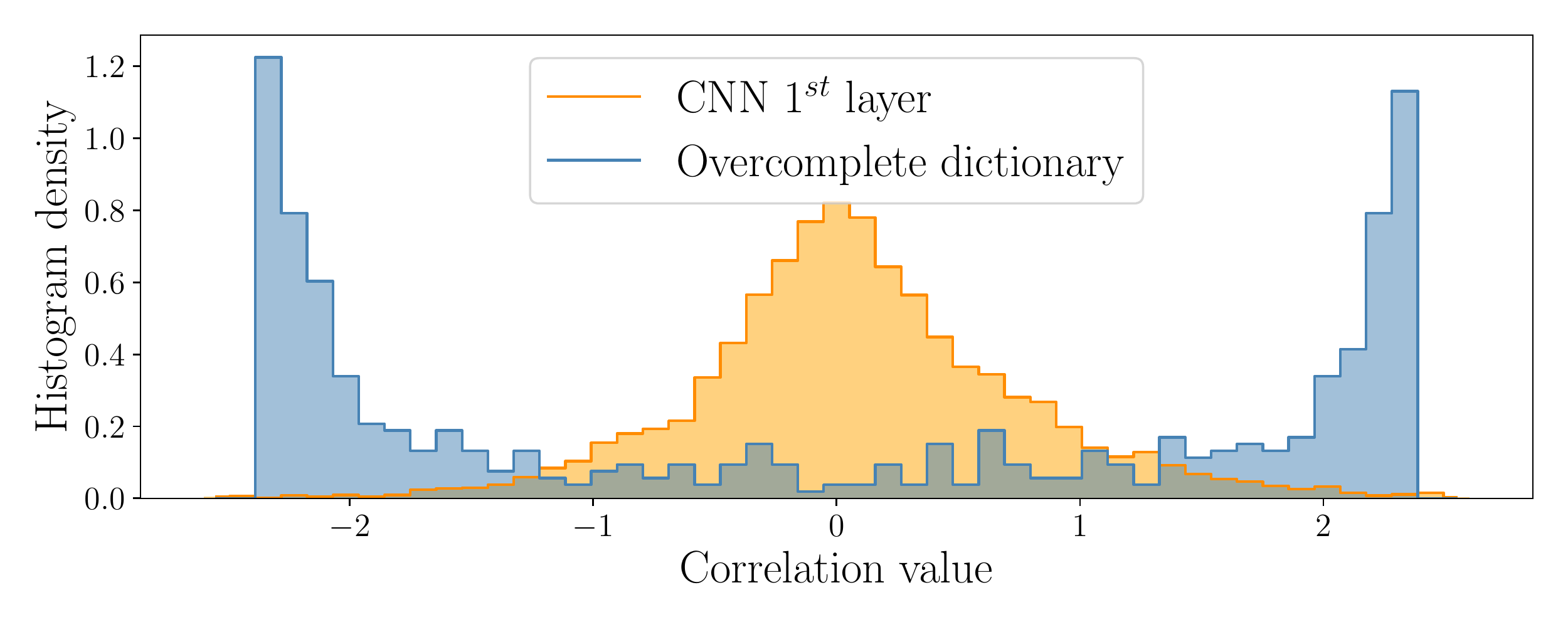}
\caption{Histogram of correlations for a typical patch with atoms of an overcomplete dictionary vs. that of activations through layer 1 filters of a standard classifier CNN.}
    \label{fig:correlations}
\end{figure}

\noindent
{\bf Rationale:} The rationale behind our encoder design is summarized as follows:
\begin{itemize}
\item An overcomplete dictionary for sparse coding results in large activations for a small fraction of the atoms, in contrast with filters learnt in the first layer of a traditional convolutional neural network where activations are clustered around zero; see Figure~\ref{fig:correlations} and Appendix.
We can therefore drop most of the activations, reducing
the effective subspace available to the attacker.
\item An attacker can still perturb the subset of top $T$ coefficients in each patch. Randomly dropping a large fraction $p$ of these coefficients allows the decoder and classifier to learn to handle randomness in the sparse code, as well as an attacker knocking a coefficient out of the top $T$.
\item The thresholds for ternary quantization of the selected coefficients are selected to provably guarantee that the attacker cannot flip the sign of any nonzero entry in the sparse code. The hard thresholding ensures that the perturbation cannot add to a coefficient which would have been selected for a clean image. Rather, the attacker must invest the effort in pushing a smaller coefficient into the top $T$, and gamble on it being randomly selected.  
\end{itemize}

\noindent
{\bf Results and Summary of Contributions:} We report on experiments on the CIFAR-10 and a subset of the ImageNet dataset (``Imagenette'') that yield interesting insights into both defense and attack strategies.
\begin{itemize}
\item 
We demonstrate the promise of a ``bottom-up'' neuro-inspired approach for design of robust neural networks that does not require adversarial training, in contrast to the top-down approaches that currently dominate adversarial machine learning. For state of the art PGD attacks, after compensating for gradient obfuscation, our adversarial accuracies are significantly better than adversarial training as in \cite{madry2017towards}.

Our results for ImageNette indicate that our patch-level sparse coding approach generalizes across image sizes.
\item
Based on experiments with a variety of attacks adapted to our defense, we come up with a novel transfer attack, based on an {\it unsupervised} version of our decoder, which reduces our adversarial accuracy to slightly below that of  \cite{madry2017towards}. This highlights the need for radically new attack strategies for novel defenses such as ours, which combine unsupervised and supervised learning.

\item
We have created our own attack library for PyTorch \cite{paszke2019pytorch}, which includes different versions of Expectation over Transformation (EOT)
for defenses utilizing stochasticity at test time, leveraging the substantial effort we have invested in attacking our defense using techniques that combat gradient obfuscation from nonlinearity and randomness. The implementation of this defense and the adversarial attack library can be found at \href{https://github.com/canbakiskan/neuro-inspired-defense}{this link}.
\end{itemize}

\section{Autoencoding Defense}\label{preprocessing}

We now discuss the details of the approach outlined in Section \ref{sec:intro}, which is illustrated in Figure~\ref{fig:block_diagram}. We discuss (standard) dictionary learning
of overcomplete representations for sparse coding at the patch level in Section \ref{sec:dictionary_learning}.  We then discuss, in Section \ref{sec:encoder}, the highly non-standard way in which we use this dictionary in the encoder. This represents the core innovation in the defense: selection of top $T$ coefficients for each patch, dropout, 
and quantization, where the quantization threshold is related to the adversarial $\ell^{\infty}$ budget that we are designing for. The CNN-based decoder,
which we describe in Section \ref{sec:decoder}, is a relatively standard architecture which restores the dimension to that of the original image, allowing us to then use a standard CNN architecture for the classifier.  However, unlike a standard autoencoder, our nominal defense is to train the decoder and classifier together in supervised fashion.
We do, however, also consider an unsupervised-trained version of the decoder, trained using a regression loss prior to supervised training of the classifier.
This provides a benchmark, but also, as we shall see, is instrumental in devising attacks adapted to our nominal defense.
Finally, the use of test-time dropout allows ensembling, as discussed in Section \ref{sec:ensemble}.

\subsection{Overcomplete Patch-Level Dictionary for Sparse Coding} \label{sec:dictionary_learning}

We consider images of size $N \times N$ with 3 RGB channels, processed using  $n \times n$ patches with stride $S$, so that we
process $M = m \times m$ patches, where $m =\lfloor{(N-n)/S}\rfloor +1$. 
Learning at the patch level allows for the extraction of sparse local features, effectively allowing reduction of the dimension of
the space over which the adversary can operate for each patch.  

We use a standard algorithm \cite{mairal2009online} (implemented in Python library \texttt{scikit-learn}), which is a variant of K-SVD \cite{elad2006image}. 
Given a set of clean training images $\mathcal{X}=\{\mathbf{X}^{(k)}\}_{k=1}^{K}$, an overcomplete dictionary $\mathbf{D}$ with $L$ atoms can be obtained by solving the following optimization problem \cite{mairal2009online}
\begin{multline}\label{eq:Dict}
\min _{\mathbf{D} \in \mathcal{C}, \{\boldsymbol{\alpha^{(k)}}\}_{k=1}^{K}} \sum_{k=1}^{K}\sum_{i,j}\Bigl( \frac{1}{2}\left\|\mathbf{R}_{ij}\mathbf{X}^{(k)}-\mathbf{D} \boldsymbol{\alpha}_{ij}^{(k)}\right\|_{2}^{2}\\ +\lambda\left\|\boldsymbol{\alpha}_{ij}^{(k)}\right\|_{1}\Bigr)
\end{multline}
where $\mathcal{C} \triangleq \bigl\{\mathbf{D} = [\mathbf{d}_{1}, \ldots, \mathbf{d}_{L}] \in \mathbb{R}^{\bar{n} \times L} \mid \left\|\mathbf{d}_{l}\right\|_{2} = 1\,, \forall l \in \{1, \ldots, L \} \bigr\}$, $\lambda$ is a regularization parameter, $\boldsymbol{\alpha}^{(k)}$ is an $m \times m \times L$ tensor containing the coefficients of the sparse decomposition, and $\mathbf{R}_{ij} \in \mathbb{R}^{\bar{n} \times \bar{N}}$ with $\bar{n} \triangleq 3n^2$ and $\bar{N} \triangleq 3N^2$ extracts the $(ij)$-th patch from image $\mathbf{X}^{(k)}$. 
The optimization problem in \eqref{eq:Dict} is not convex, but its convexity with respect to each of the two variables $\mathbf{D}$ and $\{\boldsymbol{\alpha}^{(k)}\}_{k=1}^{K}$ allows for efficient alternating minimization \cite{mairal2009online,elad2006image}.

\subsection{Sparse Randomized Encoder} \label{sec:encoder}

Based on the overcomplete dictionary obtained from \eqref{eq:Dict}, we encode the image patch by patch. For given image $\mathbf{X}$, patch $\mathbf{x}_{ij} \in \mathbb{R}^{\bar{n}}$ is extracted based on the $(ij)$-th block of $\mathbf{X}$; that is, $\mathbf{x}_{ij} = \mathbf{R}_{ij}\mathbf{X}$, and then projected onto dictionary $\mathbf{D}$ in order to obtain projection vector $\mathbf{\bar{x}}_{ij}$, where $\mathbf{\bar{x}}_{ij} = \mathbf{D}^T\mathbf{x}_{ij}$. 
Since the dictionary is highly overcomplete, a substantial fraction of coefficients typically take large values, and a sparse reconstruction of the patch can be constructed
from a small subset of these.  However, our purpose is robust image-level inference rather than patch-level reconstruction,
hence we use the dictionary to obtain a discrete sparse code for each patch using random ``population coding,'' as follows.

\noindent
{\bf 1) Top $T$ selection:} We keep only the $T$ elements of the projection vector with largest absolute values and zero out the remaining elements. The surviving coefficients are denoted by $\mathbf{\check{x}}_{ij}$.\\
{\it Rationale:} Keeping $T$ relatively large (but still a small fraction of the number of atoms $L$) provides robustness to attacks which seek to change the subset of nonzero coefficients.

\noindent
{\bf 2) Dropout:} Each of the top $T$ coefficients is dropped with probability $p$, leaving surviving outputs 
\begin{equation}
\mathbf{\tilde{x}}_{ij}(l)=\biggl\{\begin{array}{ll}
0, & \text { with probability } p \\
\mathbf{\check{x}}_{ij}(l), & \text { with probability } 1-p
\end{array},
\end{equation}
for all $l \in\{1, \ldots, L\}$. \\
{\it Rationale:} Using a large dropout probability $p$ masks the effect of an attacker ``demoting'' a coefficient from the top $T$ (the decoder and classifier are already
trained against such events). Similarly, if an attacker ``promotes'' a coefficient to the top $T$, the chances of it making it into the encoder output remain small.  

We note that the dropout used in our encoder is different from the standard use of dropout to prevent overfitting \cite{JMLR:v15:srivastava14a}.
In the latter, neurons are dropped randomly at training, but all neurons are used during testing.  In our encoder, dropout is used for both training and testing,
and is applied after the lateral inhibition corresponding to all coefficients other than the top $T$ being set to zero.

\begin{figure}[t]
    \centering
        \includegraphics[width=\linewidth]{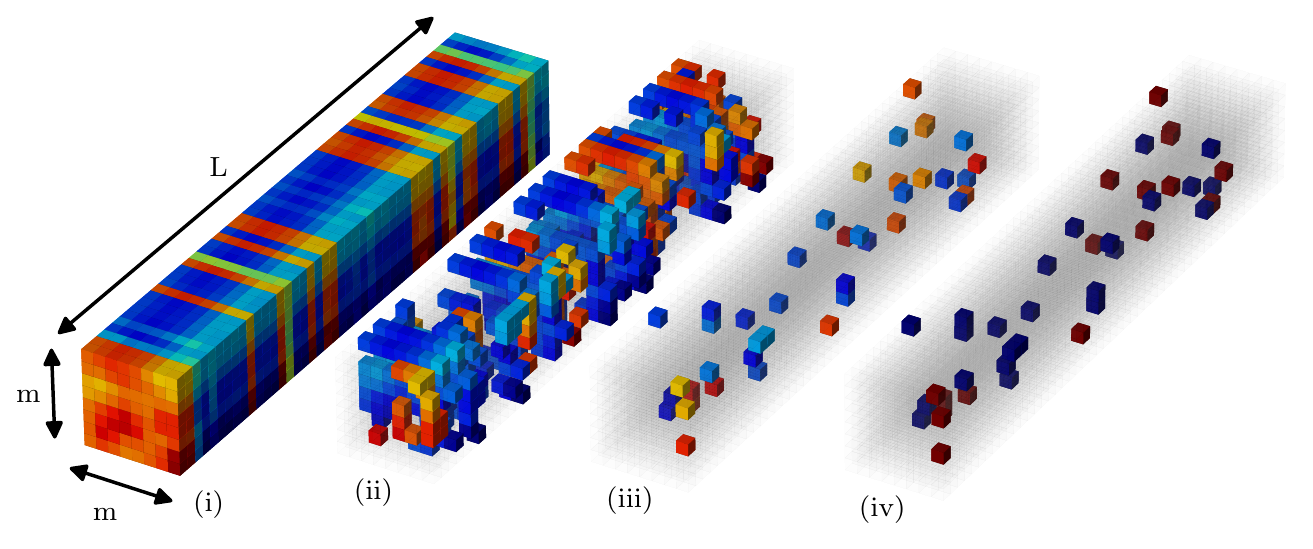}
    \caption{Progression of coefficients after each operation: (i) Each voxel shows projection onto a dictionary atom, (ii) Projections after taking top $T$, (iii) Remaining projections after dropout, (iv) Projections after activation and quantization. Notice the saturation in color.}
    \label{fig:progression}
\end{figure}

\noindent
{\bf 3) Activation/Quantization:} Finally, we obtain sparse codes with discrete values by applying binary quantization with a dead zone designed to reject perturbations.

\begin{equation}\label{eq:act_fun}
\mathbf{\hat{x}}_{ij}(l)=\biggl\{\begin{array}{ll}
\sign{(\mathbf{\tilde{x}}_{ij}(l))}\left\|\mathbf{d}_{l}\right\|_{1}, & \text { if } \frac{|\mathbf{\tilde{x}}_{ij}(l)|}{\epsilon \left\|\mathbf{d}_{l}\right\|_{1}} \geq \beta\\
0, & \text{otherwise}
\end{array},
\end{equation}
for all $l \in\{1, \ldots, L\}$, where $\beta > 1$ is a hyperparameter.\\
{\it Rationale:} By H\"{o}lder's inequality, an attacker with $\ell^{\infty}$ budget $\epsilon$ can perturb the $k$th basis coefficient by at most $\epsilon\left\|\mathbf{d}_{k}\right\|_{1}$. By choosing $\beta > 1$, we guarantee that an attacker can never change the sign of a nonzero element of the sparse code. Thus, the attacker can only demote a nonzero element to zero, or promote a zero element to a nonzero value.  As discussed, a large dropout probability alleviates the impact of both demotions and promotions.

Another consequence of choosing $\beta > 1$ is that weak patches whose top $T$ coefficients are not large enough compared to
the maximum perturbation $\epsilon\left\|\mathbf{d}_{k}\right\|_{1}$ get killed, thereby denying the adversary the opportunity to easily perturb the patch-level sparse code. Finally, the scaling of the surviving $\pm 1$ outputs by $\left\|\mathbf{d}_{l}\right\|_{1}$ acknowledges that, while the basis elements have unit $\ell^2$ norm,
their $\ell^1$ norms are allowed to vary, hence we allow basis functions whose projections survive a larger $\ell^1$ norm based threshold to contribute
more towards the decoder input.  This is entirely optional, since the decoder can easily learn the appropriate weights.

Following patch-level processing with stride $S$, the encoder outputs an image level sparse code which is an $m \times m \times L$ tensor. A typical example of how the coefficients advance through our encoder after dictionary projection is presented in Figure~\ref{fig:progression}.

\subsection{CNN-based Decoder} \label{sec:decoder}

We employ a CNN-based decoder architecture which can use redundancy across overlapping patches 
to obtain image-level information. The decoder employs three transposed convolutional layers, each followed by ReLU activation function, clipped at the
end to produce output with dimension $N \times N \times 3$ equal to that of the original RGB image.  This allows us to deploy a standard classifier
network after the decoder, and allows for a direct comparison between supervised and unsupervised decoder training.

\subsection{Ensemble Processing} \label{sec:ensemble}

In order to utilize the full potential of the randomization employed in the encoder, we allow for ensemble processing in which an input image is processed multiple (i.e., say $E$)
random realizations of our encoder at test time. 

Viewing the encoder randomization as a parameter to be averaged over, we average the softmax outputs across the $E$ realizations (see Appendix).


\section{Adversarial Attacks and Defenses}

\textbf{Attacks:} These can be broadly grouped into two categories \cite{papernot2016practical,papernot2016transferability,brendel2017decision}: whitebox attacks, in which the attacker has access to both the structure and the parameters of the neural network; and blackbox attacks, which have access only to the network outputs. Given a classifier $f: \mathbf{x} \in \mathbb{R}^N \rightarrow \mathbf{y} \in \mathbb{R}^C$, the goal of an adversary is to find a perturbation $\mathbf{e}$ that maximizes the given loss function $\mathcal{L}$ for classification under some constraints. Typically, adversarial attacks are constrained in $\ell^p$ norm, with $p=\infty$ receiving the greatest attention because it can be tuned to be imperceptible to humans \cite{goodfellow2014adversarial,kurakin2016physical,carlini2016distillationrefuted}. Among the many attack methods, Projected Gradient Descent (PGD) appears to be the most effective first order $\ell^\infty$ bounded attack, and is therefore generally used to evaluate defense methods. PGD computes the perturbation iteratively as follows:
\begin{equation}\label{eq:PGD}
\mathbf{e}_{i+1} = \text{clip}_{\epsilon}\big[\mathbf{e}_{i}+\delta \cdot \text{sign}(\nabla_\mathbf{e} \mathcal{L}(\mathbf{f}(\mathbf{x}+\mathbf{e}_i),\mathbf{y}))\big]
\end{equation}
where $\mathbf{e}_{i}$ corresponds to the value of the perturbation at iteration $i$ with $\mathbf{e}_{0} = \mathbf{0}$ or $\mathbf{e}_{0}$ with each element drawn from uniform distribution $\mathcal{U}(-\epsilon, \epsilon)$,  $\epsilon$ is the overall $\ell^\infty$ attack budget, and $\delta$ is the step size for each iteration. Expectation Over Transformation (EOT) is suggested in \cite{athalye2017synthesizing} to make attacks robust against transformations, and \cite{tramer2020adaptive} suggests using this method to evaluate defenses utilizing evaluation-time stochasticity. With EOT, PGD becomes:
\begin{equation}\label{eq:PGDwithEOT}
\mathbf{e}_{i+1} = \text{clip}_{\epsilon}\big[\mathbf{e}_{i}+\delta \cdot \text{sign}(\sum_{r=0}^{N_E-1} \nabla_\mathbf{e} \mathcal{L}_r(\mathbf{f}(\mathbf{x}+\mathbf{e}_i),\mathbf{y}))\big]
\end{equation}
where $\mathbf{e}_{0} = \mathbf{0}$ and $N_E$ corresponds to the number of multiple runs of the model. Taking the average of gradients for models utilizing randomness in evaluation time helps stabilize the gradient directions.

Reference \cite{tramer2020adaptive} motivates defense papers to extensively evaluate their defended neural networks with properly optimized threat models for the defense. Accordingly, we expend extensive effort in devising attacks optimized for our approach, incorporating EOT \cite{athalye2017synthesizing} to obtain useful gradients. \cite{obfuscated-gradients}.

\textbf{Defenses:} While there are plenty of attempts to defend against adversarial attacks \cite{madry2017towards,wong2017provable,yang2019me,buckman2018thermometer} (this is only a small subset of recent papers), the only state of the art defenses still standing are those based on adversarial training using adversarial perturbations
computed using variants of the original FGSM method \cite{goodfellow2014adversarial} of gradient ascent on a cost function: the PGD attack (iterative FGSM with random restarts) \cite{madry2017towards} is the most prominent benchmark that we compare against, but there are recent enhancements, such as the faster single-step R+FGSM scheme in \cite{wong2020fast}, and the use of a modified cost function aiming to trade off clean and adversarial accuracy (called TRADES) in \cite{zhang2019theoretically}.  In addition, substantially increasing the amount of training data using unlabeled data (and using noisy labels for these using an existing classifier) has been shown to improve the performance of adversarial training \cite{carmon2019unlabeled}. However, we do not have insight as to what properties of adversarially trained networks provide robustness, and whether these properties guarantee robustness against other attacks (conforming to the same attack budget) that have not yet been devised. Further, there is still a significant gap between clean and adversarial accuracies for an adversarially trained network for image datasets such as CIFAR-10, showing that the perturbation is not completely rejected by the network. It is claimed in \cite{schmidt2018adversarially} that this phenomenon is due to lack of data in datasets such as SVHN and CIFAR, but a more likely explanation in our view is that adversarially trained networks are still ``excessively linear.'' 

Provably robust defenses have also been studied extensively \cite{wong2017provable,croce2018provable,raghunathan2018certified}. These methods provide lower bound for adversarial accuracies; however, guarantees are provided mostly for small datasets, models, and low attack budgets. References \cite{lecuyer2019certified,cohen2019certified,salman2019provably} report certified robustness for $\ell^2$ bounded attacks which is able to scale to larger datasets such as ImageNet. Unfortunately, these certified defenses do not perform as well as adversarial training against current attack methods. Other recent defense approaches include \cite{guo2017countering,buckman2018thermometer,dhillon2018stochastic,xie2017randomization,bakiskan2020polarizing,gopalakrishnan2018robust}. However, a large number of defense methods designed based on systematic principles have been defeated by properly modified attacks \cite{athalye2017synthesizing}, or have not been shown to scale to larger datasets. 


\section{Adaptation of Attacks for Our Defense}\label{sec:adaptation}

In recent years, a majority of the proposed defense methods have been defeated by subsequent attacks \cite{tramer2020adaptive,obfuscated-gradients}, which has led to calls for each defense proposal to be evaluated not only for existing attacks, but also for attacks adapted for that particular defense \cite{onevaluating2019}.  We agree with such guidelines, and have tried a variety of whitebox and transfer attacks adapted to our defense, all of which use EOT on top of PGD to deal with the randomness in our encoder.  We have experimented extensively, and report only on the most effective attacks that we have found.

\textbf{Whitebox - Near Full Gradient Approximation (W-NFGA):} In this mode, we get close to a full whitebox attack; every operation except activation/quantization is differentiated (see Figure~\ref{fig:wnfga_attack} in Appendix). For taking top $T$ coefficients and dropout operations, the gradients are propagated to earlier layers only through nonzero coefficients. This is similar to how maxpooling operation propagates gradients. For activation/quantization, we experiment with two different backward pass approximations: in the first one we take the identity function as the approximation, in the second we consider a smooth approximation to the activation/quantization function and take the derivative of this function as the backward pass approximation (see Appendix for details). Both of the backward pass approximations result in similar adversarial accuracies.

\textbf{Whitebox - Autoencoder Identity Gradient Approximation (W-AIGA):} 
Here, for each gradient computation, the entire autoencoder is treated as having identity gradient (see Figure~\ref{fig:waiga_attack} in Appendix). 
This approximation works well only when the operation defined by the autoencoder is indeed close to identity in the forward pass, which holds for the
unsupervised-trained decoder, but not for the supervised-trained decoder which is our nominal scheme.  However, as we shall see, it is
important in devising the transfer attack described next.

\textbf{Pseudo-Whitebox - Transfer (PW-T):} In this mode, we keep the encoder dictionary fixed, but utilize an {\it unsupervised-trained decoder,} with supervised training of classifier weights as usual.  Adversarial perturbations are generated using W-NFGA or W-AIGA versions of whitebox attack, and it turns out that W-AIGA is actually more effective for when using an unsupervised-trained decoder. Our experiments show that this yields a surprisingly strong transfer attack against our defense.

\textbf{Blackbox - Transfer (B-T):} In this mode, the adversarial attack is generated based on adversarially trained classifier without taking our autoencoder model into account and then applied to our proposed autoencoder model. (See Appendix for details)

In the evaluations of our model variants including randomness, we use Expectation over Transformation (EOT) \cite{athalye2017synthesizing} to mitigate the effects of randomization as recommended in \cite{obfuscated-gradients}. Specifically, we consider PGD with EOT to evaluate the versions of our defense with randomized encoders. For deterministic encoders (considered in detailed ablation studies in supplementary materials), we consider only PGD, since EOT does not improve attack performance in these settings. Note that the attack modes W-NFGA, W-AIGA and PW-T are designed specifically for our defense, and do not apply to the benchmark defenses that we compare against.

We experiment with different variants of EOT and use the strongest one. 
We check our attack implementations by cross-testing our attacks
with the Foolbox adversarial attack toolbox, and find that the same attacks perform comparably. 


\section{Experiments, Results and Discussion}\label{results}

\newcolumntype{?}{!{\vrule \hspace{1em}}}
\newcolumntype{^}{!{\vrule \hspace{0.5em}}}
\newcolumntype{+}{!{\vrule}}
\newcolumntype{_}{!{\hspace{0.5em}}}

\subsection{Model Parameters and Settings}\label{sec:exp}

\begin{table}[!t]\centering
\begin{small}
\begin{tabular}{@{}rc@{\hspace{0.9em}}c@{\hspace{0.9em}}c@{\hspace{0.9em}}cc@{}}
\toprule
& & \multicolumn{2}{c}{\textbf{PGD with EOT}} \\ \cmidrule(lr){3-4}
& Clean & W-NFGA & W-NFGA & PW-T & B-T \\
& & Identity & Smooth &  &\\ 
\midrule
\multicolumn{1}{r^}{Our defense}& 80.06& 63.72& 61.28 & \textbf{39.53} & 57.76 \\

\bottomrule
\end{tabular}
\end{small}
\caption{Accuracies for our defense method under different attacks (CIFAR-10)}
\label{table:comprehensive_table_main}
\end{table}

Our main focus is on evaluating our defense on the CIFAR-10 dataset \cite{krizhevsky2009learning}, for which there are well-established benchmarks in adversarial ML.
This has 50000 train and 10000 test RGB images of size $32 \times 32$ ($N=32$). In order to verify that our approach scales to larger images, we also consider the Imagenette dataset: 9469 train and 3925 validation RGB images, cropped to size $160 \times 160$ ($N=160$). Both datasets contain images from 10 classes. For CIFAR-10, we use $4 \times 4$ ($n=4$) patches (i.e., $n=4$ and ambient dimension $3n^2 = 48$) and an overcomplete dictionary with $L=500$ atoms. The stride $S=2$, so the encoder output is a $15 \times 15 \times 500$ tensor ($m=15$, $L=500$). The regularization parameter in \eqref{eq:Dict} is set to $\lambda = 1$ and the number of iterations is chosen as $1000$ to ensure convergence. The hyperparameters for Imagenette are: $8 \times 8$ ($n=8$) patches and an overcomplete dictionary with $L=1000$ atoms, stride $S=4$ which gives encoder outputs of size $38 \times 38 \times 1000$ ($m=38$, $L=1000$). The regularization parameter $\lambda$ is set to $0.5$, and the number of iterations to $10000$. The guiding principle behind the choice of hyperparameters $n$ and $S$ is the empirical observation of feature sizes in relation to the size of the images. 
The number of dictionary atoms, $L$, is chosen to be $10$ times the ambient dimension for CIFAR-10, and $5$ times the ambient dimension for Imagenette,
where the limiting factor was the amount of computer memory used in dictionary learning. In order to promote sparsity, the regularization parameter $\lambda$ 
is chosen in the upper range of values that result in convergence of the dictionary learning process.

We set $T=50$, $p=0.95$. These values were found to yield the highest worst-case adversarial accuracy, based on ablation with various values of $T$ and $p$. A basis coefficient makes a nonzero contribution to the sparse code for the patch only if all three conditions are met: it is in the top $T$, it is not dropped, and it exceeds the threshold in \eqref{eq:act_fun} (we set the hyperparameter $\beta = 3$). As mentioned, we train the CNN-based decoder in supervised fashion in tandem with the classifier.
For comparison and attack design, we also consider unsupervised (US) training of the decoder. We use cross-entropy loss for supervised training.
For unsupervised decoder training, we use $\ell^2$ distance-squared as regression loss. For unsupervised training, we train the decoder for $50$ epochs. Also, in order to train the decoder, we use a cyclic learning rate scheduler \cite{smith2017cyclical} with a minimum and maximum learning rate of $\eta_{min}=0$ and $\eta_{max}=0.05$, respectively. In this scheduler, the learning rate first increases linearly in the first half of the training process and then decreases in the second half. In order to provide a consistent evaluation, we employ the ResNet-32 classifier used in \cite{madry2017towards} for CIFAR-10, and use EfficientNet-B0 \cite{tan2019efficientnet} for Imagenette. 
For supervised training (of classifier plus decoder for our nominal design, and of classifier alone for the unsupervised-trained decoder),
we use the same cyclic learning rate scheduler with the same parameters. The number of epochs is $70$ for CIFAR-10 and $100$ for Imagenette. The batch size in training is set to $64$ for both unsupervised and supervised training.

For parallel ensemble processing, after trying values in $1\leq E \leq10$, we set $E=10$, which yields the best performance, increasing clean
accuracies by up to $5\%$ (see Appendix).

For attacks, we consider PGD and PGD with EOT if it is applicable. Different from the existing EOT implementation, we use $\delta \cdot \textrm{sign}\left(\mathbf{E}_r  \left[ \nabla_x / ||\nabla_x||_2 \right]\right)$ in each step to compute the expectation, since we find in our experiments that it results in a stronger adversary.

\noindent
{\it Default attack parameters:} Unless otherwise stated, we use the following parameters for $\ell^{\infty}$ bounded PGD with EOT for CIFAR-10 trained models: an attack budget of $\epsilon = 8/255$ (as is typical in the benchmarks we consider), a step size of $\delta =  1/225$, a number of $N_S=20$ steps, a number of $N_R=1$ restarts, and a number of $N_E=40$ realizations for EOT. The same default attack parameters are used for attacking models trained on Imagenette, but given the lack of benchmarks, we test
several attack budgets $\epsilon \in \{2/255,4/255,8/255\}$.

\noindent
{\it Computation time:} On a computer with a 40-core CPU, learning the overcomplete dictionary takes 0.2 hours for CIFAR-10 and 0.8 hours for Imagenette. On a single 1080 Ti GPU, training the decoder and classifier, and computing the attack with default settings take 1.2, 1.5, and 3.5 hours, respectively for CIFAR-10. The same computations take 3, 4, and 7 hours, respectively, for Imagenette.

\noindent
{\it Benchmarks:} Our benchmarks are the PGD adversarially trained (AT) \cite{madry2017towards}, R+FGSM adversarially trained \cite{wong2020fast}, and TRADES \cite{zhang2019theoretically} defenses for the same classifier architecture. We reimplement these, to enable stress-testing these defenses with attacks of varying computational complexity. We train these models for 100 epochs with the same cyclic learning rate that we use for our models, and verify, for ResNet-32 classifier for CIFAR-10 and EfficientNet-B0 for Imagenette, that we can reproduce results obtained using
the original code . For PGD AT, training hyperparameters are $\epsilon=8/255$, $\delta=1/255$, $N_S=10$, $N_R=1$. For RFGSM AT, they are $\epsilon=8/255$, $\alpha=10/255$. For TRADES, they are $\epsilon=8/255$, $\delta=1/255$, $N_S=10$, $N_R=1$, and $\lambda_{\text{TRADES}}=1/6$.

Note that the classifier CNN used in our paper is "simple" ResNet-32 rather than the wide ResNet-32, both of which are utilized in \cite{madry2017towards} and other studies in the literature. The choice of the smaller ResNet-32 network makes evaluation of attacks computationally more feasible.

\subsection{Results, Ablation, and Discussion}

\begin{table}[!b]\centering
  \begin{small}
  \begin{tabular}{@{}rcccc@{}}
    \toprule

    & Clean  &  W-NFGA &  PW-T  & W-AIGA  \\  \midrule

    \multicolumn{1}{r_}{\underline{Our defense}} & &  &  & \\

    \multicolumn{1}{r?}{Complete} & 80.06 &  61.28 &\textbf{39.53}& 79.48  \\
    \multicolumn{1}{r?}{without A\&Q} & 81.68 & 38.48 & \textbf{37.95} & 74.05 \\
    \multicolumn{1}{r?}{without Dropout} & 76.93 & 76.61& \textbf{34.68} & 76.92 \\
    \multicolumn{1}{r?}{without Top T} & 65.72  & \textbf{23.35} & 29.95 & 59.80 \\
    \multicolumn{1}{r_}{}\\
    \multicolumn{1}{r_}{\underline{Our defense (US)}} &  & & & \\
    \multicolumn{1}{r?}{Complete} & 80.03&  65.83 & -- &\textbf{30.01} \\
    \bottomrule
  \end{tabular}
  \end{small}
  \caption{Accuracies for ablation study (CIFAR-10)}
  \label{table:ablation_table}
\end{table}

\noindent
{\bf Robustness against Defense-Adapted Attacks:}
We first investigate the performance of our defense under the different attack modes specified in Section~\ref{sec:adaptation}. Table~\ref{table:comprehensive_table_main} provides clean and adversarial accuracies for the different attack types. We note that the worst-case attack for it is {\it not} a white box attack. Rather, it is a pseudo-whitebox transfer (PW-T) attack using a network employing the same encoder but an {\it unsupervised} decoder. While this result is surprising at first, it is intuitively pleasing. An attack succeeds only to the extent to which it can change the identities of the top $T$ coefficients in the encoder.  Since the latter is designed to preserve information about the original image, providing an unsupervised decoder might provide better guidance to the attacker by giving it a reproduction of the original image to work with. This conjecture is supported by Figure~\ref{fig:unsupervised}, which shows the distribution (see Appendix for how this is computed) of the expected fraction of corrupted patches for W-NFGA and PW-T. We see that the PW-T attack results in a higher fraction of corrupted patches.

\noindent
{\bf Ablation:}
We now examine the efficacy of each component of our architecture on robustness via an ablation study in which we selectively remove one encoder component at a time and retrain the decoder and classifier, adapting attacks for each version of our defense.  We present the results of the ablation study in Table~\ref{table:ablation_table}. Both W-NFGA and W-AIGA results in Table~\ref{table:ablation_table} are obtained using the default attack parameters. For W-NFGA, we employ smooth backward pass approximation to the activation and quantization function with sharpness of $\sigma=0.25$. For PW-T, we obtain W-AIGA and W-NFGA attacks with default settings for all four different ablated {\it unsupervised} trained models and use those to test each  ablated {\it supervised} model. The reported results for PW-T are for the worst-case scenarios in which the attack achieving the lowest accuracy is considered. In all cases except the one without top $T$, the lowest accuracy for PW-T attack is obtained when W-AIGA attack is applied to the unsupervised model without A\&Q with $T=50$, $p=0.95$ and transferred. For the case without top $T$, the attack that attains minimum accuracy is W-NFGA attack obtained based on the unsupervised model without top $T$ with $p=0.95$. The overall results in Table~\ref{table:ablation_table} show that each component in our design contributes to improving robustness. The last row shows accuracies for our encoder with an unsupervised-trained decoder and separate supervised training of the classifier. We note
that our approach of joint supervised training of decoder and classifier has superior performance.

We have also conducted a detailed ablation study showing that our proposed design outperforms many other variants of our defense (see Appendix). 

\begin{figure}[!b]
  \centering
  \includegraphics[width=1.0\linewidth]{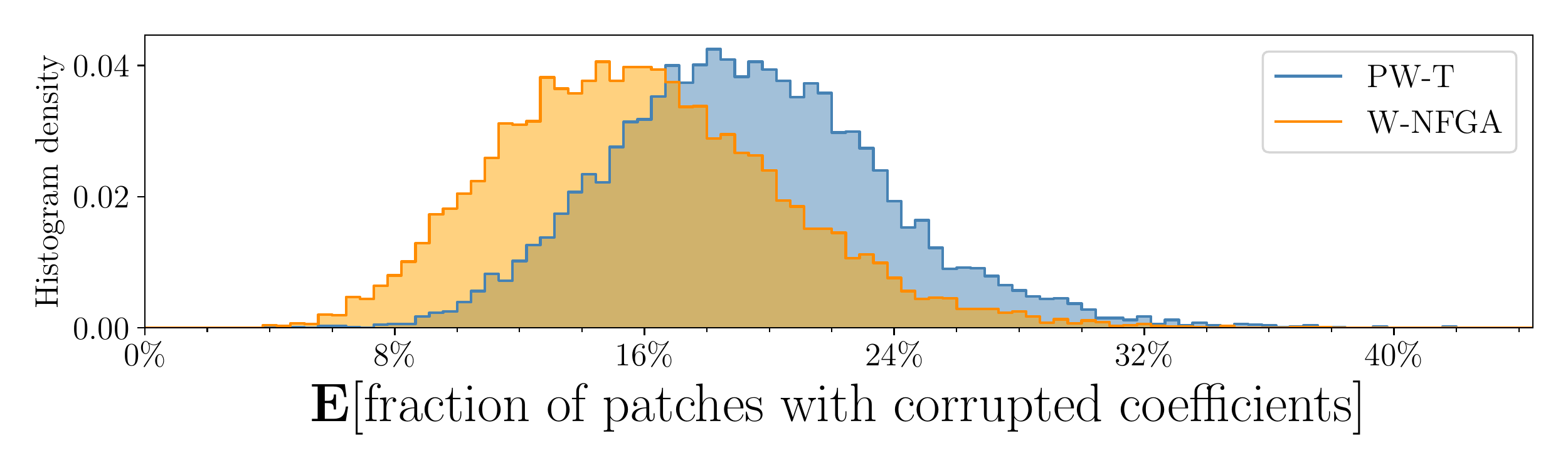}
  \caption{Histogram of $\mathbf{E}[$fraction of patches with corrupted coefficients$]$ over images for our defense under W-NFGA and PW-T attack.}
  \label{fig:unsupervised}
\end{figure}

\begin{table*}[!t]\centering
  \begin{small}
    
    \begin{tabular}{rccccc}
      \toprule
      & \multirow{3}{*}{Clean} & \multirow{3}{*}{\makecell{Adversarial\\ (Worst case)}} & \multicolumn{3}{c}{Attack Details} \\\cmidrule(lr){4-6}
      & & & \multicolumn{1}{c}{Mode}   & \multicolumn{1}{c}{Method} & \multicolumn{1}{c}{Parameters} \\

      \midrule
      \multicolumn{1}{r?}{NT} & \textbf{93.10} & 0.00 & -- & PGD & $N_S=20$, $N_R=1$ \\
      \multicolumn{1}{r?}{PGD AT \cite{madry2017towards}} & 79.41 & 42.05 & -- & PGD (C\&W Loss) & $N_S=100$, $N_R=50$ \\
      \multicolumn{1}{r?}{RFGSM AT \cite{wong2020fast}}   & 80.86 & 42.42 & -- & PGD (C\&W Loss) & $N_S=100$, $N_R=50$ \\
      \multicolumn{1}{r?}{TRADES \cite{zhang2019theoretically}}   & 75.17 & \textbf{45.79} & -- & PGD & $N_S=100$, $N_R=50$ \\
      \multicolumn{1}{r?}{Our defense} & 80.06 & 39.53 & PW-T & PGD with EOT & $N_S=20$, $N_R=1$, $N_E=40$ \\
      \bottomrule

    \end{tabular}
  \end{small}
  \caption{Comparison of our defense with other defense techniques (CIFAR-10)}
    \label{table:bottomline_table}
\end{table*}

\noindent
{\bf Comparison with benchmarks:}
We then compare our defense against naturally trained networks and our three adversarially trained benchmark defenses on the CIFAR-10 dataset. Table~\ref{table:bottomline_table} lists {\it worst-case} accuracies for each defense, where we vary the computational burden of attack on the benchmarks up to a point that is comparable to the default settings for our own EOT/PGD attack. NT denotes natural training (no defense). The worst-case adversarial accuracy for our defense is 39.53\%, which is comparable to the worst-case accuracies for the benchmarks, which range between 42-46\%. Comparing with Table~\ref{table:comprehensive_table_main}, we see that the worst-case whitebox PGD plus EOT attack for our defense is 60.28\%, about 18\% better than
the PGD attack on  \cite{madry2017towards}.

The results for the evaluation on the Imagenette dataset are given in Table~\ref{table:imagenette_table}. For NT, PGD AT, and TRADES, we use PGD attack with default parameters.

For our defense, the worst-case attack is again PW-T with transfer from the W-AIGA attack with the unsupervised-trained decoder. These experiments confirm that our defense scales to larger images. Both clean and adversarial accuracies are comparable to or exceed that of adversarial training, which is potentially also hampered by the smaller size of the training dataset.

\begin{table}[!b]\centering
  \begin{small}
  \begin{tabular}{rcccc}
    \toprule
    & \multirow{2}{*}{Clean} & \multicolumn{3}{c}{Adversarial ($\epsilon=x/255$) }\\ \cmidrule(lr){3-5}
    & & $x=2$ & $x=4$ & $x=8$ \\ \midrule
    \multicolumn{1}{r?}{NT} & \textbf{89.35} & 11.44 & 0.28 & 0.00 \\
    \multicolumn{1}{r?}{PGD AT} & 80.97 & 75.31 & 68.81 & 53.32\\
    \multicolumn{1}{r?}{TRADES} & 80.08 & 75.67 & 70.75 & 59.46\\
    \multicolumn{1}{r?}{Our defense} & 79.36 & \textbf{76.03} & \textbf{72.81}  & \textbf{65.45} \\
    \bottomrule
  \end{tabular}
  \end{small}
  \caption{Accuracies for Imagenette dataset}
  \label{table:imagenette_table}
\end{table}

\vspace*{-0.25cm}
\section{Conclusions}\label{sec:conclusions}

While our results demonstrate the potential of neuro-inspiration and bottom-up design of robust DNNs, there is significant scope for further improvement.

For example, attenuating adversarial perturbations by randomization and drastic quantization at a single step in the encoder does lead to information loss, as 
seen from the reduction in clean accuracy.  We can also visualize this information loss due to the encoder by reconstruction of the input
using the unsupervised-trained decoder, which is seen to yield less than crisp images (see Appendix).
Spreading the burden of attenuating perturbations across more network layers may help in better preserving information. 

The key to attenuating adversarial perturbations is our encoder, hence there are many possible inference architectures that can be layered on top of it. Our separation of decoder and classifier
enables reuse of standard classifier architectures, but there might be better options. Our design also enables the transfer attack from 
unsupervised-trained decoder to supervised-trained decoder, which turns out to be more effective than whitebox PGD with EOT on the original network. 
It may be more difficult to design such transfer attacks with arbitrary inference architectures, which highlights the need for 
further research on adaptive attacks, especially
for novel defenses that combine unsupervised and supervised learning, and employ concepts such as drastic nonlinearity and stochasticity.

Finally, while top-down adversarial training remains the state of the art defense, it inherits the inherent lack of interpretability and guarantees
in DNNs resulting from the curse of dimensionality for optimization in high dimensions.  

A compelling feature of a bottom-up approach
to defense is that, by focusing on attenuating perturbations over smaller segments of the input, it has the potential for evading the curse of dimensionality.

\section{Acknowledgements}
\label{sec:ack}

This work was supported in part by the Army Research Office under grant W911NF-19-1-0053, and by the National Science Foundation under grants CIF-1909320 and CNS-1518812.

\clearpage

\bibliographystyle{aaai21}

\bibliography{ms.bbl}

\clearpage


\section{Appendix}\label{sec:appendix}

\subsection{Correlations Between Dictionary Atoms and Patches}

In order to plot Figure~\ref{fig:correlations}, the normalized correlations between a given patch $\mathbf{x}_{ij}$ and dictionary atoms $\mathbf{d}_{l}$ are calculated by 

\begin{equation}
\rho_l = \frac{\langle\mathbf{x}_{ij}\,,\mathbf{d}_{l}\rangle}{\|\mathbf{d}_l\|^2}=\langle\mathbf{x}_{ij}\,,\mathbf{d}_{l}\rangle\,, \forall l \in\{1, \ldots, L\}
\end{equation}
where $\langle \cdot \,, \cdot \rangle$ represents the inner product and $\|\mathbf{d}_l\|^2=1$ for all $l \in\{1, \ldots, L\}$ by construction. The normalized activations of layer 1 filters of the standard CNN (i.e., $\mathbf{f}_p$ for $p \in\{1, \ldots, 160\}$) are calculated by 
\begin{equation}
\gamma_p = \frac{\langle\mathbf{x}_{ij}\,,\mathbf{f}_{p}\rangle}{\|\mathbf{f}_{p}\|^2}\,, \forall p \in\{1, \ldots, 160\}\,.
\end{equation}
The histograms of $\{\rho_l\}_{l=1}^{L}$ and $\{\gamma_p\}_{p=1}^{160}$ are then plotted to obtain Figure~\ref{fig:correlations}.

\begin{figure}[h]
\centering
\includegraphics[width=.99\linewidth]{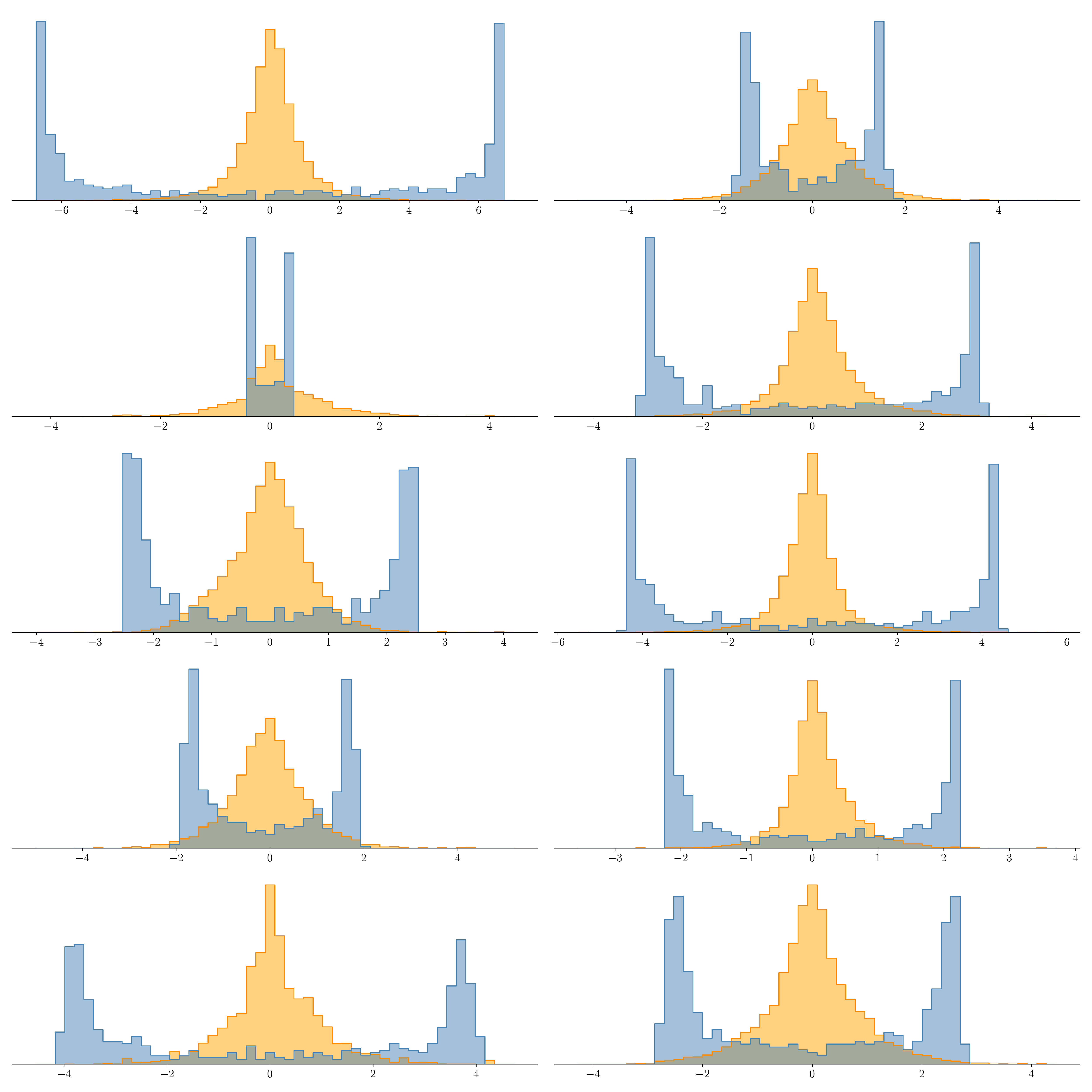}
\caption{Histograms of dictionary-atom correlations and layer 1 filter activations for 10 random patches in the style of Figure~\ref{fig:correlations}.}
\label{fig:more_correlations}
\end{figure}

In Figure~\ref{fig:more_correlations}, the histograms of correlations and activations are presented for 10 additional randomly chosen patches. As exemplified in Figure~\ref{fig:more_correlations}, most of the correlations and activations histograms exhibit the same qualitative behavior as the ``typical'' patch considered in Figure~\ref{fig:correlations}.

\subsection{Ensemble Processing}

In ensemble processing, multiple (i.e., say $E$) random realizations of our encoder are considered for an input image $\mathbf{X}$ at test time. The softmax outputs are averaged across $E$ random realizations of the autoencoder as follows:
\vspace*{-0.1cm}
\begin{equation}
h_i(\mathbf{X})=\frac{1}{E}\sum_{e=0}^{E-1}\frac{e^{g_i(\mathbf{f}_e(\mathbf{X}))}}{\sum_{j=0}^{C-1}e^{g_j(\mathbf{f}_e(\mathbf{X}))}}
\end{equation}
where $\mathbf{f}_e (\cdot)$ denotes the $e$th autoencoder realization, $g_i(\cdot)$ is the function corresponding to the $i$th class output of the classifier function $\mathbf{g}(\cdot)$, $h_i(\cdot)$ is the function corresponding to the $i$th class output for the overall ensemble model, and $C$ is the number of classes.

\begin{figure}[!t]
\centering
\includegraphics[width=.97\linewidth]{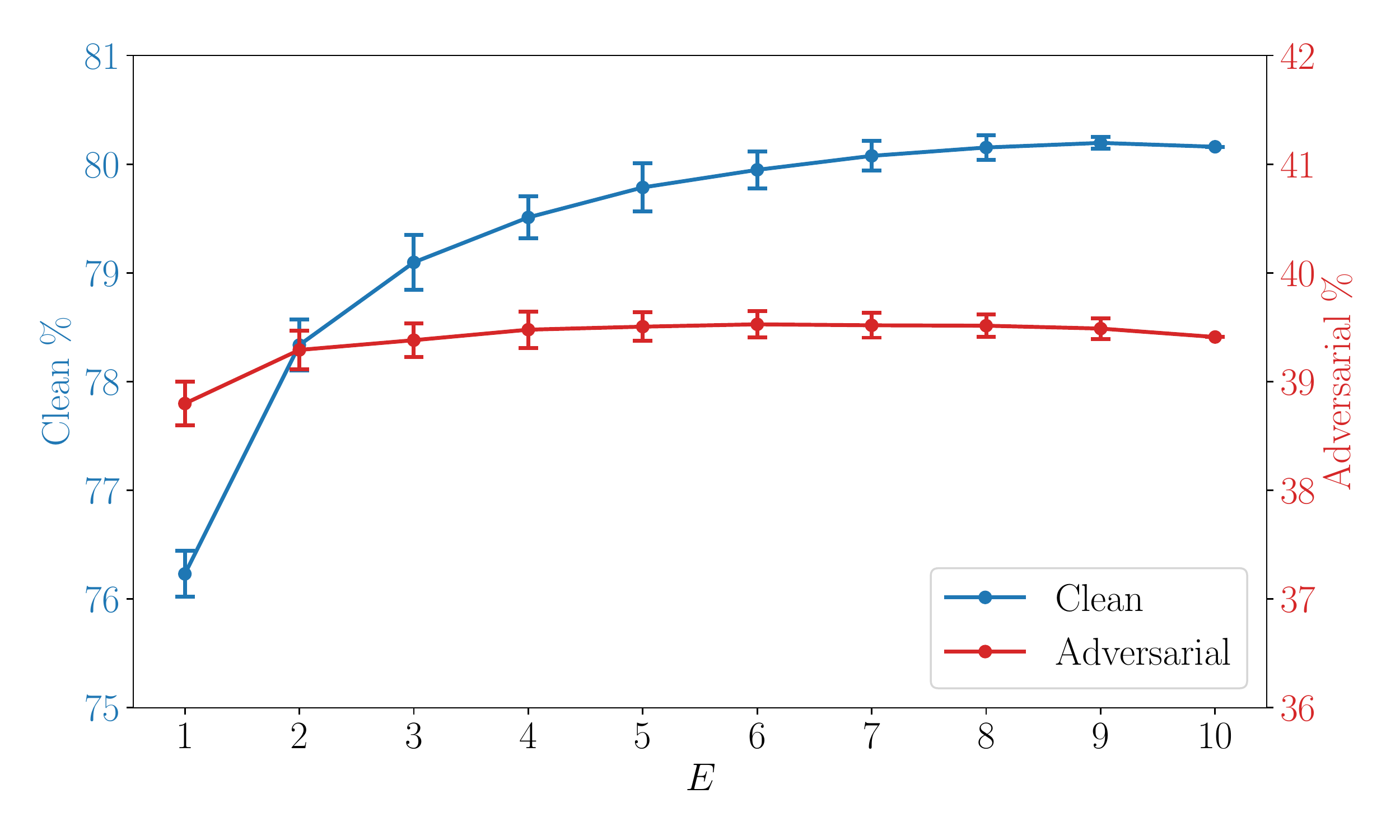}
\caption{Clean and adversarial accuracies for different number of realizations in ensemble processing.}
\label{fig:ensemble_benefit}
\end{figure}

Figure~\ref{fig:ensemble_benefit} shows the clean and adversarial accuracies for different number of encoder realizations used in ensemble processing. For our defense model, we observe that both clean and adversarial accuracies increase as the number of realizations employed in ensemble processing increases.

\subsection{Approximations to Activation/Quantization}\label{subsec:approx}

In the computation of backward pass of W-NFGA attacks, we consider two different approximations for the activation/quantization function in \eqref{eq:act_fun}: identity and smooth approximations. In the former, \eqref{eq:act_fun} is approximated as the identity function. The latter is obtained by considering a differentiable forward approximating function for \eqref{eq:act_fun} and then taking its derivative. Let $f(x)$ denote the approximate of \eqref{eq:act_fun}. The smooth backward approximation is given by
\begin{multline}
\frac{d f(x)}{dx} = \frac{\|\mathbf{d}_k\|_1}{2\sigma} \biggl(\sech^2\Bigl(\frac{x - \beta\epsilon\|\mathbf{d}_k\|_1}{\sigma}\Bigr) \\+ \sech^2\Bigl(\frac{x + \beta\epsilon\|\mathbf{d}_k\|_1}{\sigma}\Bigr)\biggr)
\end{multline}
where 
\begin{multline}
f(x) = \frac{\|\mathbf{d}_k\|_1}{2}\biggl(\tanh\Bigl(\frac{x-\beta\epsilon\|\mathbf{d}_k\|_1}{\sigma}\Bigr)\\+\tanh\Bigl(\frac{x+\beta\epsilon\|\mathbf{d}_k\|_1}{\sigma}\Bigr)\biggr)
\end{multline}
with $\sigma$ determining the sharpness of the approximation.

\subsection{Attack Mode Details}
\begin{figure}[!!h]
  \centering
  \includegraphics[width=.9\linewidth]{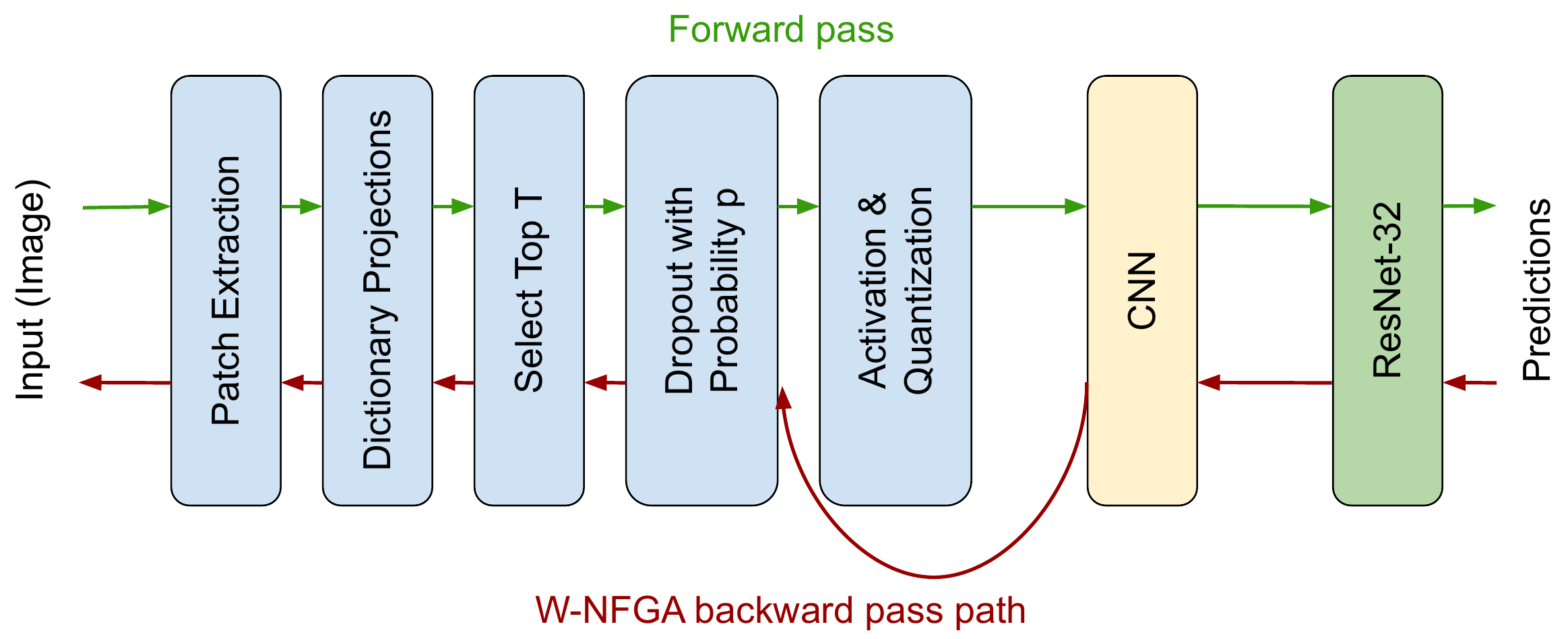}
  \caption{W-NFGA attack forward/backward pass}
  \label{fig:wnfga_attack}
\end{figure}

In Figure~\ref{fig:wnfga_attack}, the backward and forward passes of W-NFGA attack mode are shown. In the forward pass, all components of the defense are employed. In the backward pass, the gradients of all components except Activation/Quantization are calculated and used. In order to calculate the gradients of the activation/quantization function, we consider identity and smooth approximations as explained in Section~\ref{subsec:approx}.

\begin{figure}[!h]
  \centering
  \includegraphics[width=.9\linewidth]{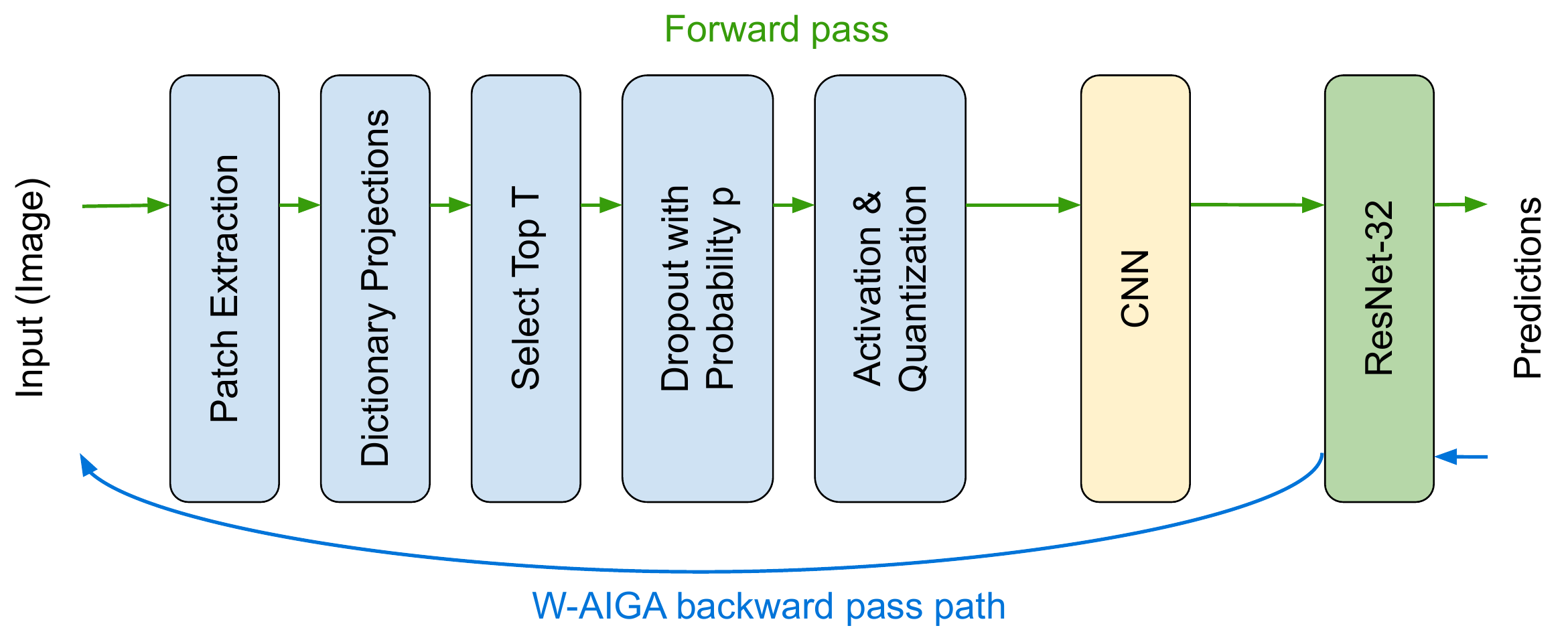}
  \caption{W-AIGA attack forward/backward pass}
  \label{fig:waiga_attack}
\end{figure}

Figure~\ref{fig:waiga_attack} shows the forward and backward passes used in W-AIGA attack mode. In the forward pass, all components of the defense are considered. In the backward pass, only the gradients of the classifier are used and carried to the input layer in each step of the attack by bypassing the autoencoder.

For blackbox transfer attacks (B-T), we use the attacks generated with default attack parameters for PGD adversarially trained model of \cite{madry2017towards}. 

\subsection{Computation of the Histograms in Figure~\ref{fig:unsupervised}}

For a given image with index $i$, let $q_i(j)$ represent the probability that a total of $j$ coefficients in top $T$ are ``toppled'' by the attack before dropout. $q_i(j)$ is found empirically by taking the histogram of number of ``toppled'' coefficients over patches for the image $i$.

We can then calculate the probability of a randomly selected patch in image $i$ {\it not} being corrupted by the attack as
$$
\zeta_i = \sum_{j=0}^{T} q_i(j)\,p^j
$$
where $p$ is the probability of dropping a coefficient (Dropout rate).

We note that $\zeta_i$ is simply the probability that coefficients that were not in the top $T$ in the clean image $i$ for this patch are dropped by the dropout mechanism. Of course, conditioned on this ``no corruption'' event, the sparse code for the patch has a different distribution from that of the clean image because of the coefficients which have been knocked out of the top $T$. But it still belongs to the original ensemble of possible sparse codes for that patch for a clean image. We ignore these subtleties, and compute the expected fraction of patches whose sparse code has at least one nonzero coefficient which would not have appeared in the clean image as $z_i= 1-\zeta_i$. The histogram of $z_i$ across all images is what is plotted in Figure~\ref{fig:unsupervised}.

\begin{figure*}[!t]
\centering
\includegraphics[width=.97\linewidth]{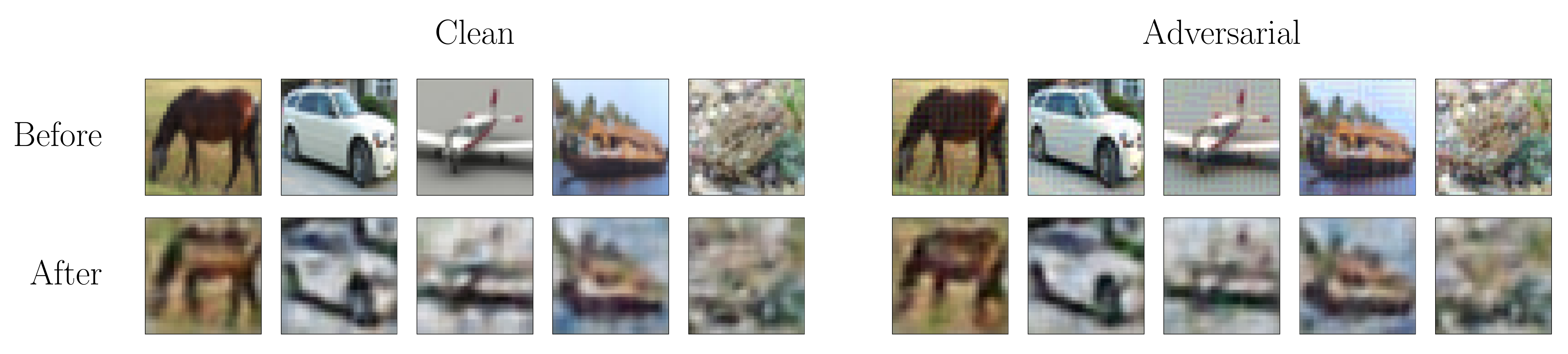}
\caption{Clean and adversarial images before and after our autoencoder based on unsupervised-trained decoder.}
\label{fig:before_after}
\end{figure*}

\subsection{Additional Ablation Studies}\label{subsec:appendix_ablation}

In order to better assess the contributions played by the components of our defense, we have tested variations of our defense with different hyperparameters and architectures.
While the encoder in our nominal defense with parameter settings $T=50,p=0.95$ yields the best attacked accuracy, we find that our results are remarkably
resilient to the specific values of $T$ and $p$, as long as the expected number of selected coefficients, $T(1-p)$, is roughly the same. We report here on three models that use the same architecture as our defense but with different hyperparameters, namely, models with ($T=1, p=0.00$), ($T=5, p=0.50$), and ($T=10, p=0.75$). These hyperparameters are chosen to have the expected value of surviving coefficients ($T(1-p)$) within the same order of magnitude as our original defense. 
As reported in Table~\ref{table:ablation_extended_table}, these hyperparameters all yield similar results.
Intriguingly, the adversarial accuracy for a deterministic scheme with no dropout ($T=1, p=0.00$) is closest to the robustness of our nominal defense ($T=50, p=0.95$). However, for a version of our defense with unsupervised decoder training, randomness is
far more important: the attacked accuracy (under the worst-case attack) is $30.01\%$ for our nominal encoder ($T=50, p=0.95$) versus $19.64\%$ for the deterministic encoder ($T = 1, p=0.00$). This highlights the need for further research into how best to optimize the sparse codes produced by our encoder architecture.

\begin{table}[!htb]\centering
  \begin{small}
    \begin{tabular}{rccc}
    \toprule

    & Clean & W-NFGA &  PW-T   \\  \midrule

    \multicolumn{1}{r_}{\underline{Our defense}} & & &\\
    \multicolumn{1}{r?}{Nominal : $T=50$, $p=0.95$} & 80.06 & 61.28 & \textbf{39.53}\\
    \multicolumn{1}{r?}{$T=1$, $p=0.00$} & 81.48  & 65.82 & \textbf{39.42}\\
    \multicolumn{1}{r?}{$T=5$, $p=0.5$}  & 85.26  & 67.70 & \textbf{38.63}\\
    \multicolumn{1}{r?}{$T=10$, $p=0.75$}  & 84.78 & 64.14 & \textbf{36.40}\\
    \multicolumn{1}{r_}{}\\[-1ex] 
    \multicolumn{1}{r_}{\underline{Other architectures}} & & &\\
    \multicolumn{1}{r?}{Sparse Autoencoder}  & 91.33  & \textbf{0.06}  & 71.88 \\
    \multicolumn{1}{r?}{Classifier w/ dropout} & 88.22  & \textbf{0.08} & 66.05\\
    \multicolumn{1}{r?}{Gaussian blur prepr.}   & 91.59  & \textbf{0.00} & 72.80 \\
    \bottomrule
  \end{tabular}
  \end{small}
  \caption{Accuracies for additional ablation studies in CIFAR-10}
  \label{table:ablation_extended_table}
\end{table}

We also report on our experiments with three architectures which incorporate different aspects
of our defense. In the first, we use a k-sparse autoencoder \cite{makhzani2013k} trained in supervised fashion. It uses $k=50$ out of 500 channels in the bottleneck layer. In the second architecture, we expand the number of channels in the first layer of the standard classifier we use in our earlier experiments and apply dropout (at inference time) with $p=0.95$ to this layer. This model does not differ from the standard classifier in any other way. In the third architecture, based on the observation in Figure~\ref{fig:before_after} that our frontend results in a blurred image, we apply Gaussian blurring to the images before they go into the classifier. The standard deviation of the Gaussian blur filter is determined by minimizing the $\ell^2$ loss between Gaussian blurred images and images processed by unsupervised version of our defense and set to $\sigma_G=0.625$.  As seen from the results presented in Table~\ref{table:ablation_extended_table},
attacked accuracies drop to zero for all of these architectures, indicating that the robustness of our proposed architecture results from the unique combination of ideas incorporated in our defense.

\subsection{Analysis of Computational Budget}\label{subsec:appendix_computation}

In Table~\ref{table:bottomline_table}, we aim to keep the computational budget comparable across the attacks to our and other defense methods. We can estimate the computational budget $C$ as $C=K\times N_S\times N_R \times N_E$ where $K$ is the computational cost of a single backward pass, which depends on the overall model size. For our defense, the cost of a backward pass is $K = K_{ours}$, which is 5.4 times the backward pass cost with $K = K_{AT}$ for PGD AT, RFGSM AT, and TRADES; that is, $K_{ours} = 5.4\times\,K_{AT}$. This is mainly due to the large number of filters used in our decoder structure. In Table~\ref{table:bottomline_table}, PGD AT, RFGSM AT, and TRADES take $C=5\times10^{3}\times K_{AT}$ computation steps whereas the default attack to our defense takes $C=8\times10^{2}\times K_{ours}$, which is roughly the same computational budget.

Since attacking our defense is so computationally intensive, we had used the following settings for the attacks on our defense in Table~\ref{table:bottomline_table}: $N_S=20,N_R=1, N_E=40$.  These parameters were optimized so as to keep the computational budget to within an order of magnitude of those typically needed to effectively attack adversarially trained networks.  In Table~\ref{table:computation_table}, we report on adversarial accuracies for our defense as 
we increase the computational budget further, by increasing each of the parameters in turn, keeping the others fixed.  The results
in Table~\ref{table:computation_table} confirm that increasing the computational budget 
relative to the ``default'' attack reported on in Table~\ref{table:bottomline_table} 
does not decrease the adversarial accuracies for our defense significantly.

\begin{table}[!h]\centering
  \begin{small}
    \begin{tabular}{rc@{\hspace{0.9em}}c@{\hspace{0.9em}}c@{\hspace{0.9em}}c}
    \toprule

    & \multicolumn{4}{c}{\textbf{PW-T}}   \\   \cmidrule(lr){2-5}
    & Default & $N_S=100$ &  $N_E=100$ & $N_R=10$  \\  \midrule
    \multicolumn{1}{r?}{Our defense} & 39.53 & 38.93 & 39.25 & 39.38 \\
    \bottomrule
  \end{tabular}
  \end{small}
  \caption{Accuracies for different computational budget attacks to our defense in CIFAR-10}
  \label{table:computation_table}
\end{table}

\subsection{Images Before and After Autoencoding}

While our nominal defense employs a supervised-trained decoder, we can visualize the information loss due to our randomized sparse coding strategy using an unsupervised decoder trained based on reconstruction loss. We present in Figure~\ref{fig:before_after} clean and adversarial images, before and after the autoencoder, for this unsupervised-trained version of our defense model.
The encoder is as for our nominal defense, with $T = 50$, $p=0.95$.

We note that there is an appreciable degradation in sharpness due to the autoencoder, implying that there is significant scope for improvement in our design, despite the promising gains in robustness that have been demonstrated.



\end{document}